\documentclass{article}
\usepackage{iclr2023_conference,times}

\usepackage{microtype}

\usepackage{amsmath}
\usepackage{amssymb}
\usepackage{mathtools}
\usepackage{amsthm}
\usepackage{graphicx}
\usepackage{booktabs}
\usepackage{amsmath}
\usepackage{amssymb}
\usepackage{amsfonts}
\usepackage{multirow}
\usepackage{verbatim}
\usepackage{caption}
\usepackage{supertabular}
\usepackage{float}
\usepackage{amsmath}
\usepackage{amsmath}
\usepackage{tabularx} % Prompt table
\usepackage{listings, listings-rust} % code
\lstdefinelanguage{JavaScript}{
  morekeywords=[1]{break, continue, delete, else, for, function, if, in,
    new, return, this, typeof, var, void, while, with},
  morekeywords=[2]{false, null, true, boolean, number, undefined,
    Array, Boolean, Date, Math, Number, String, Object},
  morekeywords=[3]{eval, parseInt, parseFloat, escape, unescape},
  sensitive,
  morecomment=[s]{/*}{*/},
  morecomment=[l]//,
  morecomment=[s]{/**}{*/}, % JavaDoc style comments
  morestring=[b]',
  morestring=[b]"
}[keywords, comments, strings]

\usepackage{enumitem}
\usepackage{xcolor}
\usepackage{xspace}
\usepackage{textcomp}
\usepackage{makecell}
\usepackage{multicol}
\usepackage{lscape} 
\usepackage{siunitx}
\usepackage[linesnumbered,ruled,vlined]{algorithm2e}
\usepackage{algorithmic}

\usepackage{amssymb}% http://ctan.org/pkg/amssymb
\usepackage{pifont}% http://ctan.org/pkg/pifont
\usepackage{scrextend}

\usepackage{array}
\usepackage{latexsym}
\usepackage[T1]{fontenc}
\usepackage[utf8]{inputenc}
\usepackage{microtype}
\usepackage{xurl} % URL typesetting with line breaks
\usepackage{nicefrac}       % compact symbols for 1/2, etc.
\usepackage{changepage}
\usepackage{xargs}
\usepackage{wrapfig,lipsum,booktabs}
\usepackage{longtable}
\usepackage{subcaption}
\usepackage{pgfplots}
\usetikzlibrary{pgfplots.groupplots}
\pgfplotsset{compat=1.3}
\usepackage{tikz}
\usetikzlibrary{patterns}
\usepackage{placeins} % For appendix to make sure floats are all in their respective sections (FloatBarrier)

\usepackage[most]{tcolorbox}

\usepackage{hyperref}
\usepackage[capitalize,noabbrev]{cleveref}
\crefname{section}{Section}{\S\S}
\crefname{section}{Section}{\S\S}
\crefname{table}{Table}{Tables}
\crefname{figure}{Figure}{Figures}
\crefname{algorithm}{Algorithm}{}
\crefname{equation}{eq.}{}
\crefname{appendix}{Appendix}{}
\crefformat{section}{Section #2#1#3}

\definecolor{mydarkblue}{rgb}{0,0.08,0.45}
\hypersetup{
    colorlinks=true,
    linkcolor=mydarkblue,
    filecolor=blue,      
    urlcolor=mydarkblue,
    citecolor=violet
}

\newcommand{\papertitle}[0]{\vspace{-2mm}Differentiation of Multi-objective Data-driven Decision Pipeline}

\DeclareSymbolFont{extraup}{U}{zavm}{m}{n}
\DeclareMathSymbol{\varheart}{\mathalpha}{extraup}{86}
\DeclareMathSymbol{\vardiamond}{\mathalpha}{extraup}{87}

\iclrfinalcopy

\usepackage[capitalize,noabbrev]{cleveref}

\theoremstyle{plain}

\theoremstyle{definition}

\theoremstyle{remark}

\usepackage[loose]{minitoc} % parttoc

\noptcrule

\makeatletter
\AtBeginDocument{%
  \renewcommand{\sectionautorefname}{\S\@gobble}

  \renewcommand{\subsectionautorefname}{\S\@gobble}  
}
\makeatother

\title{\center Differentiation of Multi-objective Data-driven Decision Pipeline}

\author{%
   Peng Li\\
  Cainiao Network \\
  \texttt{lipeng.lipeng@cainiao.com} \\
  \And
  Lixia Wu\thanks{Corresponding author}\\
  Cainiao Network \\
  \texttt{wallace.wulx@cainiao.com} \\
  \And
  Chaoqun Feng\\
  Cainiao Network \\
  \texttt{fengchaoqun.fcq@cainiao.com} \\
  \And
  Haoyuan Hu\\
  Cainiao Network\\
  \texttt{haoyuan.huhy@cainiao.com} \\
  \And
  Lei Fu\\
  Cainiao Network\\
  \texttt{leo.ful@cainiao.com} \\
  \And
  Jieping Ye\\
  Alibaba DAMO Academy \\
  \texttt{yejieping.ye@alibaba-inc.com} 
  }

\fancypagestyle{firstpage}{
  \lhead{Differentiation of Multi-objective Data-driven Decision Pipeline}
  \rhead{\normalsize{\today}}
}

\begin{document}

\doparttoc % Tell to minitoc to generate a toc for the parts
\faketableofcontents % Run a fake tableofcontents command for the partocs

\maketitle
\thispagestyle{empty}

%% ANON
%\begin{tikzpicture}[remember picture,overlay,shift={(current page.north west)}]
%\node[anchor=north west,xshift=17cm,yshift=-1.5cm]{\scalebox{0.7}[0.7]{\includegraphics[width=5.5cm]{figures/octopack.png}}};
%\end{tikzpicture}

% ANON
%\vspace{-14mm}
\begin{abstract}
Real-world scenarios frequently involve multi-objective data-driven optimization problems, characterized by unknown problem coefficients and multiple conflicting objectives. Traditional two-stage methods independently apply a machine learning model to estimate problem coefficients, followed by invoking a solver to tackle the predicted optimization problem. The independent use of optimization solvers and prediction models may lead to suboptimal performance due to mismatches between their objectives. Recent efforts have focused on end-to-end training of predictive models that use decision loss derived from the downstream optimization problem. However, these methods have primarily focused on single-objective optimization problems, thus limiting their applicability. We aim to propose a multi-objective decision-focused approach to address this gap. In order to better align with the inherent properties of multi-objective optimization problems, we propose a set of novel loss functions. These loss functions are designed to capture the discrepancies between predicted and true decision problems, considering solution space, objective space, and decision quality, named landscape loss, Pareto set loss, and decision loss, respectively. Our experimental results demonstrate that our proposed method significantly outperforms traditional two-stage methods and most current decision-focused methods.
\end{abstract}

%\begin{morekeywords}
%decision-focused learning, multi-objective optimization, smart prediction-and-optimization, data-driven optimization
%\end{morekeywords}

\section{Introduction}
Uncertain decision-making is prevalent in various real-life scenarios, such as personalized recommendations\cite{Liu23} and path planning \cite{XuL0X023} based on route time prediction. These scenarios involve a workflow for handling data-driven decision problems where parameter coefficients are predicted based on environmental or historical information, and decisions are made using these predictions. For instance, in recommendation systems, click-through rate prediction and sorting or top-K recommendation based on click-through rates are common examples. However, most traditional approaches decompose this workflow into two independent phases: the prediction phase and the decision phase. Obtaining a perfect prediction model is often unachievable. The problem coefficients generated by the prediction model are frequently noisy. Given that conventional prediction models prioritize predictive accuracy, they often neglect the structure and attributes of downstream optimization problems. However, in most of optimization problems, the impact of problem coefficients on the final decision is not uniform. Without a decision-focused approach, decisions derived from imperfectly predicted coefficients are more prone to deviate from the optimal solution. Consequently, the separation of the prediction and decision phases often leads to suboptimal outcomes due to the misalignment of objectives between these phases.\cite{Amos17}. Addressing this issue, the integration of prediction models with decision problems for unified training has emerged as a promising direction across various domains \cite{Kotary22,Vlastelica20}.

%Moreover, obtaining a perfect prediction model is often unachievable. Without further decision-focused trick, decisions based on unperfected predicted coefficients often deviate from the true optimal solution.
%This separation often results in suboptimal performance due to the mismatch between the objectives of the prediction and decision phases\cite{Amos17}. 
%
% The separation between the prediction and decision phases can result in poorer performance due to objective mismatches. Addressing this issue, the integration of prediction models with decision problems for unified training has emerged as a promising direction across various domains \cite{Kotary22,Vlastelica20}.

The integration of optimization problems into contemporary deep learning frameworks poses significant challenges, particularly because the mapping from predicted coefficients to the optimal decision variable is non-differentiable. To address the aforementioned issue, numerous decision-focused methods have been proposed to train a predictive model by minimizing decision loss associated with  the downstream optimization task \cite{Kotary21}. Considering the diverse forms and complexity of real-world optimization problems, as well as the time-intensive aspect of iterative problem-solving, prior research has concentrated on two principal directions. The first direction seeks to broaden the applicability of the approach by formulating additional optimization problems of varied forms, employing differentiable operators \cite{Amos17,Wilder18}. The second direction \cite{Shah22} entails the development of efficient decision surrogate functions to diminish training time or amplify decision-focused performance. However, previous studies have predominantly concentrated on the domain of single-objective decision problems, with less emphasis on extending the learning paradigms into the domain of multi-objective optimization.

Numerous data-driven multi-objective problems exist in real-world scenarios \cite{Gunantara18, Hua21}. For example, online resource allocation and display advertisement optimization represent classic instances of data-driven multi-objective optimization problems. Such problems necessitate considering the interests of the advertising platform, users, and advertisers. Their objectives typically encompass enhancing user shopping experiences, fulfilling advertiser goals, and augmenting matching efficiency alongside revenue maximization for the advertising platform. Multi-objective optimization demonstrate increased complexity in the search space compared to their single-objective counterparts. Optimal solutions for multi-objective problems are frequently non-unique and correspond to a Pareto front. Additionally, the gradient vectors of distinct objectives often exhibit conflicting directions. Given the challenges' significance and complexity, we aim to propose a novel Multi-Objective Decision-Focused Learning (MoDFL) model to address these gaps. To the best of our knowledge, this represents the inaugural effort to apply the decision-focused paradigm to the realm of multi-objective optimization.
%Considering the difficulty and significance of addressing these challenges, our objective is to propose a novel multi-objective decision-focused learning (MoDFL) model to fill the aforementioned gap.

To address the mentioned inquiries, we propose three decision-focused loss functions tailored for multi-objective optimization problems. The proposed loss function involves three modules, which measure the distance of objective space, solution space as well as the decision quality of representative point. Specifically, we introduce a landscape loss based on the sample rank maximum mean discrepancy (sRMMD) to quantify the discrepancy in the objective space across optimization problems. The objective space is considered a manifold within high-dimensional space, represented by the set of objective vectors corresponding to the solutions. We propose a Pareto set loss to directly measure the distance within the Pareto set, aiming to circumvent the homogeneity that may impede model training in certain optimization problems. For instance, \begin{math} f^1(\pi)= 2\pi^2-4\pi\end{math} and \begin{math}f^2(\pi)= \pi^2-2\pi\end{math} exhibit identical landscapes and share the same optimal solution. The  decision loss is analogous to other decision-focused losses, utilizing the decision quality of a representative solution as the loss criterion. This representative solution is derived from employing the weighted sum method in this paper. Differentiation of the proposed loss function is achieved by reparameterization method and transforming the multi-objective problem into a corresponding single-objective optimization problem. Building on these modules, MoDFL integrates the predictive model and multi-objective optimization problem into a unified pipeline, which enables end-to-end training of the system.

%
%
%
%
% Extensive experiment exhibit our proposed MoDFL have competitive performance compared to two-stage method and the state-of-art method.

The remainder of this paper is structured as follows. Section II provides an overview of the related work. Section III details the formulation of the data-driven multi-objective problem and introduces pertinent definitions. Section IV provides a motivating example on multi-objective decision learning. In Section V, we present three novel loss functions, explore the differentiation of multi-objective optimization mapping, and explain the integration of the aforementioned modules within MoDFL. The performance of MoDFL is compared to that of two-stage methods and other state-of-the-art DFL method in Section VI. Finally, we conclude this paper in Section VII.

\section{related work}

\subsection{Differentiating Optimization Problems as a Layer in Neural Network}
A plethora of studies has investigated the integration of prediction problems with downstream decision-making processes. The scope of this study includes methodologies such as Smart Predict-and-Optimize (SPO)\cite{Elmachtoub22}, End-to-End Optimization Learning\cite{Kotary21}, and Decision-Focused Learning (DFL)\cite{Mandi21}. The central aspect of this topic aims to the differentiable mapping from problem coefficients to optimal decision variables, a concept we denote as the decision gradient herein. Utilizing the decision gradient enables the formulation of many optimization problems as differentiable operators within gradient-based methods like neural networks.
% This domain encompasses methodologies named smart predict-and-optimize (SPO)\cite{Elmachtoub22}, end-to-end optimization learning\cite{Kotary21}, and decision-focused learning (DFL)\cite{Mandi21}, all under the purview of this investigation

%This research area encompasses methods such as smart predict-and-optimize (SPO), end-to-end optimization learning, and decision-focused learning, all of which are within the scope of investigation. 
%
%The central aspect of this topic revolves around the differentiation of the mapping from problem coefficients to optimal decision variables, which we refer to as the decision gradient in this section. By employing the decision gradient, it becomes possible to design most optimization problems as differentiable operators within gradient-based methods, such as neural networks.
%the calculation of the gradient of the optimal decision variable in relation to the predicted problem coefficients, 
	Much of the seminal literature in this field stems from the work of Amos \cite{Amos17}. This approach proposed in \cite{Amos17} leveraged the Karush-Kuhn-Tucker (KKT) optimality conditions for quadratic programming and the implicit function theorem to construct decision gradients. Nevertheless, this method's limitation lies in its requisite for a full-rank Hessian matrix of the objective function, constraining its applicability. Subsequent research has addressed this constraint by employing elaborate techniques to generalize the methodology across a broader spectrum of optimization problems. For instance, Wilder \cite{Wilder18} introduced one quadratic regularization term into the objective function to extend the method proposed by Amos \cite{Amos17} to linear programming (LP). Ferber et al. \cite{Ferber20} extended decision-focused learning to mixed-integer programming (MIP), relaxing MIP to LP by applying a cutting plane at the root LP node and leveraging Wilder's method \cite{Wilder18} to enable end-to-end training. Xie et al. \cite{Xie20} approached the top-K problem through the lens of the optimal transport problem, extending Mandi and Guns's work \cite{Mandi20} to tackle it. Additionally, the aforementioned analytical differentiation and analytical smoothing  of optimization mappings also have been successfully applied to various applications, including maximum satisfiability problem \cite{Wang19-1}, model predictive control \cite{Amos18-2}, game theory \cite{Ling18}.
%A significant portion of the relevant literature originates from the groundbreaking work by Amos \cite{Amos17}. 
		Another category of methods has emerged from the smart predict-and-optimize (SPO) method \cite{Elmachtoub22}. This approach primarily focuses on optimization problems with linear objectives and involves predicting the parameters of the optimization problem. It introduces a convex surrogate upper bound on the loss, facilitating an accessible subgradient method. Vlastelica et al. \cite{Vlastelica20} also explore linear objective problems, computing decision gradients by perturbing the predicted problem coefficients. Expanding on this, Niepert et al. \cite{Niepert21} improve the method by incorporating noise perturbations into perturbation-based implicit differentiation to maximize the posterior distribution. 
	
	Notably, all these methods require iterative addressing optimization problems during training, incurring substantial computational costs. To mitigate this issue, Mulamba et al. \cite{Mulamba20} implemented a solution cache to record solutions discovered during model training and devised a surrogate decision loss function based on contrastive loss to reduce the time spent solving optimization problems. Kong et al. \cite{Kong22} employed an energy-based model to characterize decision loss, thereby reducing the overhead in end-to-end training by linking minimum energy with minimum decision loss. Shah et al. \cite{Shah22} introduced the surrogate convex loss functions such as WeightedMSE, Quadratic, DirectedWeightedMSE, and DirectedQuadratic, to alleviate the computational burden. Mandi et al. \cite{Mandi21} leveraged the learning-to-rank concept to devise a proxy for decision loss and introduced four surrogate decision loss functions.

\subsection{Data-driven Multi-Objective Optimization}

Multi-objective optimization, the simultaneous optimization of multiple objective functions, is prevalent in practical applications such as computational advertising \cite{Liu23}, database retrieval \cite{Zheng2022MO}, and numerical simulation of physical experiments \cite{Jin2009}. A myriad of methods, including gradient-based techniques such as the Multiple-Gradient Descent Algorithm (MGDA) \cite{Desideri12} and population-based approaches like MOEA/D \cite{Zhang07}, have been proposed to solve the multi-objective problems.  Furthermore, multi-objective optimization has been successfully applied across various domains within the machine learning community, including kernel learning \cite{Li14}, sequential decision-making \cite{Roijers13}, multi-agent systems \cite{Roijers13, Parisi14}, and multi-task learning \cite{Sener18, Ma20, Lin19}. More details on multi-objective problems could refer to surveys such as \cite{Gunantara18,Hua21}.

However, the prevailing assumption in most research is that the parameters of optimization problems are fixed and precise. This assumption marks a significant difference between data-driven multi-objective problems and conventional optimization problems. Bayesian optimization \cite{Shah16} and surrogate-assisted evolutionary computation \cite{jin2018data} are two typical approaches for addressing multi-objective data-driven optimization problems, aside from DFL. These methods are designed for optimization problems that often lack explicit representation and whose solution evaluation can be time-consuming, such as the optimization of hyper-parameters for machine learning models. Employing suitable surrogate models to approximate the objective space can significantly reduce the runtime of optimization algorithms and improve the search efficiency. Conversely, DFL assumes a known and constant problem structure, with variability confined to problem parameters that require estimation through predictive models. However, it is noteworthy that both approaches still construct the loss functions of predictive models on the basis of predictive accuracy rather than the ultimate decision loss. This constitutes the main difference between Bayesian optimization and Surrogate-assisted evolutionary computation in comparison to DFL.

%Bayesian optimization \cite{Shah16}
%
%More details on multi-objective problem could refer to surveys such as \cite{Gunantara18,Hua21}. While many decision-focused learning models are designed for predictive models based on neural networks, they are not suitable for solving large-scale problems or training neural networks. Therefore, our focus is on gradient-based multi-objective optimization techniques. These methods aim to find a descent direction that simultaneously reduces all objectives by leveraging multi-objective Karush-Kuhn-Tucker  conditions \cite{Kuhn13}. Gradient-based approaches have been successfully applied in various domains, such as kernel learning \cite{Li14}, sequential decision making \cite{Roijers13}, multi-agent learning \cite{Roijers13, Parisi14}, Bayesian optimization \cite{Shah16}, and multi-task learning \cite{Sener18, Ma20, Lin19}.
%comprehensive overview of this field, readers can refer to surveys such as
\begin{figure*}[htbp]
\centerline{\includegraphics[width=\columnwidth]{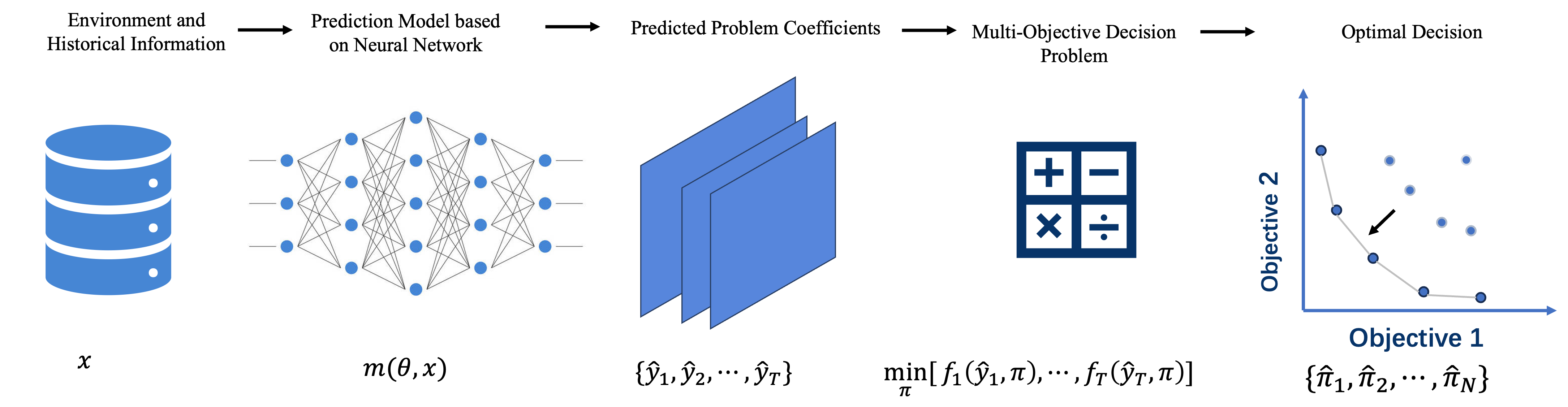}}
\caption{The pipeline of data-driven multi-objective problem}
\label{fig1}
\end{figure*}

\section{problem description}
This study concentrates on decision-focused learning within the context of multi-objective optimization problems. The pipeline of addressing the data-driven decision problem is illustrated in Fig.\ref{fig1}. The given data include the contextual information (features) \begin{math}x_i\end{math}, the true coefficients \begin{math}y_i\end{math}, as well as the formulation of a parametric multi-objective optimization problem. The subscript \begin{math} i \end{math} denotes the index of each optimization problem instance. In the decision-focused setting, the coefficients \begin{math} y_i \end{math} are not known in advance but can be estimated with a machine learning model. The procedure is divided into two phases: prediction and optimization. The prediction phase involves estimating the problem coefficients based on \begin{math} x_i \end{math}; The optimization phase need to solve the optimization problem where the coefficients are fixed as predicted result. The goal is  to optimize the objective values of the solution obtained in the optimization phase, using the true problem coefficients.
%includegraphics[width=1.0\columnwidth]{myimage.png}

During the prediction phase, the prediction model is represented as \begin{math} m(\theta, x) \end{math}, and the predicted problem coefficients can be defined as \begin{math} \hat{y_i} \equiv m(\theta, x_i) \end{math}. During the optimization phase, the multi-objective optimization problems can be formulated as presented in Eq.\ref{eq:(1)}.  
\begin{equation} \label{eq:(1)}
  \begin{aligned}
& \underset{\pi}{\text{min}}
& & f(y,\pi) = [f_1(y^1,\pi),\cdots,f_T(y^T,\pi)] \\
& \text{s.t.}
& & g(y,\pi) \leq 0 \\
\end{aligned}
\end{equation}
Where  \begin{math} \pi \end{math} denotes one feasible solution,  \begin{math} f(\cdot) \end{math} denotes the \begin{math} T \end{math} objective functions, and \begin{math} g(\cdot) \end{math} denotes the constraint functions. With the predicted problem coefficients, its optimal solution is defined as  \ \begin{math}  \hat{\pi_i}(\theta, x_i) \equiv \arg \underset{\pi,g(\hat{y_i},\pi) \leq 0}{\operatorname{min}} f(\hat{y_i},\pi) \end{math}. The problem under study can be formulated as follows:

\begin{equation} \label{eq:(2)}
 \underset{\theta}{\text{min}}
 \ L(x,y,\theta) = \mathbb{E}_{x_i,y_i\in \mathbb{D}}[f_1(y^1_i, \hat{\pi_i}),\cdots,f_T(y^T_i,\hat{\pi_i}) ]
\end{equation}
where \begin{math} \mathbb{D}  \end{math} denotes the latent distribution of input data. 

In Eq.\ref{eq:(2)}, the objective functions refer to a vector rather than a scalar, distinguishing it from single-objective optimization problems. Additional definitions pertinent to multi-objective optimization problems are introduced. 

\textbf{Pareto Dominance}: Pareto dominance is established when, for two vectors \begin{math} \phi= [\phi_1,\phi_2,\cdots,\phi_T] \end{math} and \begin{math} \psi= [\psi_1,\psi_2,\cdots,\psi_T] \end{math}, it holds that \begin{math} \phi_i \leq \psi_i \end{math}  for all \begin{math} i=1,2,\cdots,T \end{math}, and there exists at least one index \begin{math} i \end{math} for which \begin{math} \phi_i < \psi_i \end{math}. In such cases, \begin{math} \phi \end{math} is said to dominate \begin{math} \psi \end{math}.

Pareto dominance provides a means to compare the quality of two solutions within multi-objective optimization problems. This concept facilitates the introduction of optimal solution definitions in multi-objective optimization, namely, Pareto optimality and the Pareto optimal set.

\textbf{Pareto optimality}: If and only if there does not exist another solution \begin{math}  \pi  \end{math} so that \begin{math}  f(\pi^{*}) \end{math} is dominated by \begin{math}  f(\pi) \end{math}, the solution \begin{math} \pi^{*} \end{math}  is said to be Pareto optimal.

The set of all Pareto optimal solutions for a multi-objective optimization problem is termed the Pareto optimal set \begin{math} PS^{*} \end{math}, and the corresponding objective vectors constitute the Pareto front.

\textbf{Pareto front}: Given a multi-objective optimization problem and its Pareto optimal set \begin{math} PS^{*} \end{math}, the Pareto front (\begin{math}PF^{*} \end{math}) is defined as the set 
\begin{math} PF^{*} = \{f(\pi) = (f^1(\pi), ..., f^T(\pi))|\pi \in PS^{*}\} \end{math}.

\section{Motivating Example}
In the realm of multi-objective optimization, conflicting objectives can exacerbate the disjunction between predictive models and downstream optimization problem, thereby diminishing the efficacy of the performance. To elucidate this phenomenon, we offer an illustrative example within this subsection.

Consider a simple bi-objective optimization scenario represented by \begin{math} f^1(\pi)= a_1\pi^2-a_2\pi; f^2(\pi)= a_1\pi^2-a_3\pi\end{math}, where \begin{math} a_1 (a_1>0)\end{math} is a predefined constant, and \begin{math} a_2, a_3\end{math} represent the true coefficients of the problem, which are estimated through a machine learning algorithm. We consider a condition wherein the predictive model's test loss fails to converge upon zero. Precisely, there exists a positive scalar \begin{math} \varepsilon\end{math} satisfying the condition \begin{math} |\hat{a_i}- a_i |\geq \varepsilon, i=2,3\end{math}, with \begin{math} \hat{a_2}, \hat{a_3}\end{math} denoting the coefficients as forecasted by the model. Furthermore, it is postulated that the model achieves an admissible level of precision, namely, \begin{math} |\hat{a_i}- a_i | \leq |\frac{a_2-a_3}{2}| , i=2,3 \end{math}. The Pareto set of true optimization problem \begin{math} PS^{*} \end{math} is the interval lying between \begin{math} \frac{a_2}{2a_1} \end{math} and \begin{math} \frac{a_3}{2a_1} \end{math}. For the predicted coefficients, the corresponding Pareto set \begin{math} \hat{PS^{*}} \end{math} is the interval between \begin{math} \frac {\hat{a_2}}{2a_1} \end{math} and \begin{math} \frac{\hat{a_3}}{2a_1} \end{math}. It can be inferred that the maximal length of the intersecting interval between these Pareto set is \begin{math} \frac{|a_2-a_3|- \varepsilon}{2a_1} \end{math}. As the dimensionality \begin{math} n\end{math} of \begin{math} \pi\end{math} increases, the proportion of the overlap \begin{math} \hat{PS^{*}}  \cap  PS^{*} \end{math} relative to the area of \begin{math} PS^{*} \end{math}, which is \begin{math} (1-\frac{\varepsilon}{|a_2-a_3|})^n \end{math}, will diminish to zero. Furthermore, increasing the number of objectives in the optimization problem invariably results in a progressively diminished ratio of \begin{math} \hat{PS^{*} } \cap PS^{*} \end{math} to \begin{math} PS^{*}  \end{math}.

In the context of decision-focused learning with multiple objectives, the two-stage approach is challenged when the loss function of the predictive model does not converge to zero. The decisions derived from imperfectly predicted coefficients are more prone to deviate from the optimal solution. It is noteworthy that quadratic decision problems, despite being among the most elementary, yield explicit optimal solutions.  For more complex optimization problem, the complex functional space and the necessity to balance between objectives pose significant hurdles in identifying the mapping between optimal solutions and problem coefficients and investigate the differentiation of the mapping.  Thus, it is imperative to delve into the interplay between learning algorithms and downstream optimization problems and to integrate these aspects into a unified framework.

\section{Methodology}

In this section, we delineate the integration of multi-objective optimization and  predictive models based on neural network to facilitate decision-focused training.  Decision-focused learning encompasses two primary elements: the decision surrogate loss function and the differentiation of optimization mappings. We will elaborate on these elements in the context of multi-objective decision-focused learning and detail the implementation of our proposed methodology.

\subsection{Decision surrogate loss}

The essence of decision surrogate loss involves measuring the divergence between the predicted optimization problem and the true optimization problem, taking into account the structural aspects of the optimization problem.  Considering the complexity of the decision space in multi-objective optimization and the potential conflicts among objectives, we have developed loss functions that capture the intricacies of multi-objective optimization problems. These functions are designed to measure the discrepancies in the objective space, solution space, and representative points between the predicted problem and the true problem.

\subsubsection{\textbf{Landscape loss}}

In single-objective optimization problems, the objective space is one-dimensional, where a partial ordering can characterize the relationship between the objective function values of two solutions. However, in multi-objective optimization, the T-dimensional objective space renders the partial order relationship inadequate for representing the variations in objective space across different problems. Moreover, the stringent criteria for establishing Pareto dominance often result in numerous objective vectors lacking any dominance relationship, especially for many-objective optimization problems.

To address these limitations, we analogize the multi-dimensional objective space to the manifold in high-dimensional space, such as images, video, or audio signals.  We use the concept of neighborhood relations in high-dimensional spaces to clearly define how different sets of objectives are connected and to identify common patterns in their overall distribution. Specifically, we utilize the objective vector of solutions found during training model to approximately represent the objective space, and employ the sample rank maximum mean discrepancy (sRMMD) \cite{Shoaib2023}  to measure the distance between the objective spaces of different optimization problems. This proposed metric is referred to as the landscape loss function.

To develop the concept of sRMMD, we initially discuss entropy-regularized optimal transport, which involves determining an optimal “coupling” \begin{math} c\end{math}  between a source distribution \begin{math} P \end{math} and a target distribution \begin{math} Q \end{math}.   The \begin{math} \prod{(P,Q)} \end{math} represents the set of joint probability measures on the product space with marginal distributions \begin{math} P\end{math} and \begin{math} Q \end{math}, where \begin{math} \phi \in P\end{math} and \begin{math} \psi \in Q \end{math}. The entropy-regularized optimal transport problem is formulated as follows:

 \begin{equation} \min _ {c \in \prod{(P,Q)}}   \int  \frac {1}{2} |\phi-  \psi|^ {2} \, d  c  (\phi,  \psi)+  \varepsilon  KL(  \pi  ||P  \otimes  Q)
\end{equation}
, where the Kullback–Leibler (KL) divergence is defined by:
\begin{equation}  KL(  \pi  ||P  \otimes  Q) = \int  \log \frac{d  c  (\phi,  \psi)}{dP(\phi)\,dQ(\psi)} d  c  (\phi,  \psi) \end{equation}

The dual form of the previously described problem is provided in Eq.\ref{eq:(5)}, with its derivation detailed in \cite{Genevay2019}.
\begin{equation} 
\label{eq:(5)}
\begin{split}
  \max_ {u,v}  & \int   u(\phi)dP(\phi)+\int  v(\psi)dQ(\psi)+  \varepsilon   -  \\
  & \varepsilon  \int   \int   exp \left[  \frac {1}{\varepsilon }  (u(\phi)+v(\psi)-  \frac {1}{2}  \left|\phi-\psi \right|^ {2}  ) \right] dP(\phi)dQ(\psi)
\end{split}
\end{equation}

where the maximization is over the pairs \begin{math} u \in L_1(P), v \in L_1(Q)  \end{math}. The optimal entropic potentials for \begin{math} \varepsilon \end{math} are the pair of functions  \begin{math} u_{\varepsilon}  \end{math} and \begin{math} v_{\varepsilon} \end{math}  correspond to the functions that achieve the maximum in  Eq.\ref{eq:(5)}. This dual formulation can be solved using methods such as gradient descent or Sinkhorn's algorithm \cite{Cuturi2013}. By executing a predetermined number of iterations and applying automatic differentiation in PyTorch or TensorFlow, we can compute the gradient of sRMMD. Building upon \begin{math} u_{\varepsilon} \end{math} and \begin{math} v_{\varepsilon} \end{math}, we present the soft rank map in its sample-based form as follows:

\begin{equation} 
R^{n}_{\varepsilon}(\phi) = \frac{\sum^{N}_{i=1} \psi_ {i} exp(\frac {1}{\varepsilon } (v^n_{\varepsilon}(\psi_i)-\frac {1}{2}\left|\phi-\psi_ {i}\right|^ {2}))
 }{\sum^{N}_{i=1} exp(\frac {1}{\varepsilon } (v^n_{\varepsilon}(\psi_i)-\frac {1}{2}\left|\phi-\psi_ {i}\right|^ {2})}
\end{equation}

For a distribution \begin{math} P  \end{math} , the sample rank map \begin{math} \mathbb{R}^{m} \end{math} is defined as the plug-in estimate of the transport map from \begin{math}  P\end{math} to a uniform distribution \begin{math}Q = Unif([0, 1]^d)\end{math}, where \begin{math}  d\end{math} denotes the dimensionality. 

%\frac {exp(\frac {1}{\varepsilon }(g_ {e}^ {n}Y_ {i})-\frac {1}{2}|x-Y_ {i}|^ {2}}{x_ {p}(\frac {1}{\varepsilon }(g_ {e}(Y_ {i})-\frac {1}{2}|x-Y_ {i|}^ {2}}
%\begin{split}
%  \max_ {m,n}  & \int   m(\phi)dP(\phi)+\int  n(\psi)dQ(\psi)+  \varepsilon   -  \\
%  & \varepsilon  \int   \int   exp \left[  \frac {1}{\varepsilon }  (m(\phi)+n(\psi)-  \frac {1}{2}  \left|\phi-\psi \right|^ {2}  ) \right] dP(\phi)dQ(\psi)
%\end{split}

Let \begin{math}k\end{math} : \begin{math}\mathbb{R}^{m} \times \mathbb{R}^{m} \rightarrow \mathbb{R}\end{math} be a characteristic kernel function. Let  \begin{math} R_{\tau,\varepsilon}(\phi)\end{math} denote the soft rank map of \begin{math} P_{\tau}\end{math} for \begin{math}\tau \in (0,1)\end{math}. The distribution \begin{math} P_{\tau}\end{math} is the mixture distribution of \begin{math}P_{X}\end{math} and \begin{math}P_{Y} \end{math}, where \begin{math} P_{\tau} = \tau P_{X} +(1-\tau)P_{Y}\end{math}. Samples \begin{math}X_i,X_j \end{math}  are drawn from \begin{math}P_{X} \end{math}; similarly, samples  \begin{math}Y_i,Y_j \end{math} are drawn from \begin{math}P_{Y} \end{math}. The sRMMD between the distribution of \begin{math}P_{X}\end{math} and \begin{math}P_{Y} \end{math} can is expressed as follows:
% Let \begin{math}P_{X}\end{math} and \begin{math}P_{Y}\end{math} be two i.i.d. probability measures and let \begin{math} X,X^{'} \sim  P_{X}\end{math} and \begin{math} Y,Y^{'} \sim  P_{Y}\end{math}. 
% \begin{equation} 
%R^{n}_{\varepsilon}(\phi) = \frac{\sum^{N}_{i=1} \psi_ {i} exp(\frac {1}{\varepsilon } (v^n_{\varepsilon}(\psi_i)-\frac {1}{2}\left|\phi-\psi_ {i}\right|^ {2}))
% }{\sum^{N}_{i=1} exp(\frac {1}{\varepsilon } (v^n_{\varepsilon}(\psi_i)-\frac {1}{2}\left|\phi-\psi_ {i}\right|^ {2})}
%\end{equation}

\begin{equation} 
\begin{split}
\mathrm{sRMMD}_{\tau,\varepsilon}^{m,n}(P_{X},& P_{Y})^{2}  \cong {\frac{1}{m^{2}}}\sum_{i,j=1}^{m}k(\mathrm{R}_{\tau,\varepsilon}^{m+n}(X_{i}),\mathrm{R}_{\tau,\varepsilon}^{m+n}(X_{j})) \\
& +{\frac{1}{n^{2}}}\sum_{i,j=1}^{m}k(\mathrm{R}_{\tau,\varepsilon}^{m+n}(Y_{i}),\mathrm{R}_{\tau,\varepsilon}^{m+n}(Y_{j}))- \\
&  {\frac{2}{nm}}\sum_{i=1}^{m}\sum_{j=1}^{n}k(\mathrm{R}_{\tau,\varepsilon}^{m+n}(X_{i}),\mathrm{R}_{\tau,\varepsilon}^{m+n}(Y_{j}))
\end{split}
\end{equation}
%where \begin{math} x^{'} \end{math} is an independent copy of x with the same distribution, and \begin{math} y^{'} \end{math} is an independent copy of y.

In multi-objective decision focused learning, we consider the objective vector corresponding to a particular solution as a sample within a high-dimensional space. Assume we have a pooling of solutions \begin{math} S \end{math}, where  \begin{math} \pi_i \in S \end{math}. The objective vectors for the predicted and true problems can be represented as \begin{math} \phi_i, \psi_i \end{math}, respectively, where \begin{math} \phi_i,  \psi_i \in \mathbb{R}^{|S|\times T}  \end{math} and \begin{math} i \end{math} denotes the index of the optimization problem instance

\begin{equation} 
\begin{split}
L_{l}(x, y, \theta) &=  {\frac{1}{N}}\sum_{i}\mathrm{sRMMD}_{\tau,\varepsilon}^{\left|S\right|,\left|S\right|}(\phi_i,\psi_i) \\
&=  {\frac{1}{N}}\sum_{i}\mathrm{sRMMD}_{\tau,\varepsilon}^{\left|S\right|,\left|S\right|}({f(y,\pi_i)},{f(m(\theta,x),\pi_i)}) 
\end{split}
\end{equation}

We employ a 6-dimensional Gaussian mixture kernel as our kernel function, as described in \cite{Shoaib2023}, with bandwidth parameters \begin{math} \sigma = (1,2,4,8, 16,32)  \end{math}. The associated hyperparameters, \begin{math}\tau=0.5, \varepsilon=10^{-5}\end{math}. For additional implementation details, readers are directed to \cite{Shoaib2023}.
% , are consistent with those in \cite{Shoaib2023}.

\subsubsection{\textbf{Pareto set loss}} 
Optimal solutions may differ across various optimization problems. However, certain optimization problems, despite having different coefficients, can exhibit identical landscapes and share the same optimal solution. For example, a normalization for the coefficients of LP can preserve both the optimization landscape and solution optimality. In light of this homogeneity, we introduce a loss function that directly measures the distance between solution spaces of optimization problems. With a focus on optimization problems that target optimal solutions, our proposed loss function quantifies the disparity between sets of optimal solutions. Drawing on the concept of inverted generational distance, this loss function employs the minimum distance between the Pareto sets of the predicted and true problems in solution space. Let \begin{math} PS^{*}_y \end{math} represent the Pareto set of the true problem and \begin{math} PS^{*}_{\hat{y}} \end{math} denote the Pareto set of the predicted problem. The proposed loss function is formulated as follows:

\begin{equation} 
\label{eq:(9)}       
L_{ps}(y,\hat{y},w) =  {\frac{1}{N}}\sum_{i}d(\hat{\pi},PS^{*}_{y}) \end{equation}

where \begin{math} d(\cdot)\end{math} refers to the Euclidean distance. Pareto sets typically manifest as hyperplanes or intervals. Considering computational complexity, we approximate the actual Pareto set using a finite set of representative points. Consequently, \begin{math} d(\hat{\pi},PS^{*}_{y})  \end{math} is approximated by \begin{math} min_{\pi^{*}_{y}\in PS^{*}_{y}} d(\hat{\pi},\pi^{*}_{y})  \end{math}. Using this approximation, the loss function in Eq. \ref{eq:(9)} is reformulated as:

\begin{equation}        
L_{ps}(x, y, \theta) =  {\frac{1}{N}}\sum_{i}min_{\pi^{*}_{y_i}\in PS^{*}_{y_i}} d(\hat{\pi},\pi^{*}_{y_i}) \end{equation}  
Given that it quantifies the divergence between the Pareto sets of distinct optimization problems, the loss function is referred to as the Pareto set loss function.

\subsubsection{\textbf{Decision loss}}
Similar to other single-objective decision-focused learning approaches, our approach adopt the decision quality associated with the Pareto optimal solutions of the predicted problem as the loss function. However, due to the multidimensional nature of the Pareto front, the aforementioned loss function cannot be directly applied in the multi-objective scenario. Here, we transform the predicted problem and the true problem into single-objective problems using a certain method. Subsequently, we employ the optimal solutions of the transformed predicted problem in the objective function of the transformed true problem as the loss function, which is named as decision loss function.

We apply the conventional weighted sum method to transform multi-objective problem into single-objective problem. Due to the difference in the magnitude of different objectives, the predicted coefficients are initially processed through an instance normalization layer. The process is detailed as follows:
\begin{equation}
 BN(\hat{y^j_i}) =  \frac{\hat{y^j_i}-mean(\hat{y^j_i})}{std(\hat{y^j_i})}; j= 1,\cdots,T
 \end{equation}
 where  \begin{math} mean(\hat{y^t_i}) \end{math} and   \begin{math} std(\hat{y^t_i}) \end{math} denote the mean and standard deviation of \begin{math} \hat{y^t_i} \end{math} respectively. In optimization problems with LP and MIP, it's easy to prove that the instance normalization layer preserves the relative cost value ordering. Consequently, the normalization layer maintains the optimization landscape and the optimality of solutions unchanged.

Equipped with uniform weight, we focused the weighted optimization problem \begin{math}  f_{w}(\hat{y},\pi)= \frac{\sum_{j}f^j(BN(\hat{y^j}),\pi)}{T} \end{math}. This method ensures that the mapping from problem coefficients to the optimal solution aligns with that of a single-objective decision-focused method. Let \begin{math} \hat{\pi} \end{math} denote as the optimal solution of \begin{math}  f_{w}(\hat{y},\pi)\end{math}. The decision loss is then articulated as follows: 
\begin{equation}    
\label{eq:(11)} 
L_{d}(x, y, \theta) =  \frac{\sum_{i}\sum_{j}f(BN(y^j_i),\hat{\pi}_i)}{NT} \end{equation}   

\subsection{Differentiation of optimization mappings}
Differentiation of optimization mappings refers to the process of computing the gradient of the decision surrogate loss with respect to the predicted problem coefficients. As previously discussed, we can observe that the decision surrogate loss is a function of \begin{math} y \end{math}, \begin{math} \hat{y} \end{math}, and \begin{math} \hat{\pi} \end{math}. 
When \begin{math} \hat{\pi} \end{math} is sampled from the given distribution of solution (or solution cache), the gradient of the decision surrogate loss on the search space loss function can be calculated using reparameterization techniques \cite{Kingma15}. Accordingly, the gradient of the landscape loss function is determined in this manner.

For the gradients of Pareto set loss function and decision loss function, the decision gradient of optimization mappings is decomposed into two terms by the chain rule, as expressed below:
\begin{equation}
 \frac{\partial L(x, y, \theta)}{\partial \hat{y} } =   \frac{\partial L(x, y,\theta)}{\partial  \hat{\pi} (\theta, x)} \frac{\partial \hat{\pi} (\theta, x)}{\partial  \hat{y} } 
 \end{equation}

 The first term denotes the gradients of the decision loss with respect to the decision variable. In this work, the proposed decision loss function are both the continuous function on the decision variable, facilitating effortless automatic differentiation by deep learning frameworks.

 The second term corresponds to the gradients of the optimal decision with respect to the predicted coefficient. This term involves the non-differentiable \begin{math} argmin \end{math} operator. To overcome this issue, various efficient surrogate functions and carefully designed techniques have been proposed. With the prevalent use of linear programming in practical applications, we demonstrate our approach through data-driven linear programming. The differentiation of the optimal condition for smooth linear programming (DSLP) \cite{Wilder18}  is utilized to derive the gradient of the optimal decision relative to the problem coefficients.
 We illustrate our methodology using data-driven linear programming as an example. Notably, our approach extends to various problem types, provided that a suitable differentiable operator \cite{Mandi2023} is substituted for DSLP. For a clearer exposition of DSLP, we first delineate linear programming with the subsequent equations:

\begin{equation}
\label{eq:(14)}
  \begin{aligned}
& \underset{\pi}{\text{min}}
& & f(y,\pi) = [y^1\pi,\cdots,y^T\pi] \\
& \text{s.t.}
& & A\pi \leq b \\
\end{aligned}
\end{equation}
% Linear programming is a mathematical optimization technique used to find the best possible solution to a problem, given a set of linear constraints and an objective function.
By leveraging the KKT conditions, the DSLP constructs a system of linear equations based on the predicted coefficient and the optimal decision. Applying the implicit function theorem allows us to derive the expression for the second term.
\begin{equation}
 \quad
\begin{bmatrix} 
\nabla_{\hat{\pi}}^2 f(y,\hat{\pi})
& A^T \\
\ \emph{diag}(\lambda)A &  \emph{diag}(A \hat{\pi}-b)
\end{bmatrix}
\begin{bmatrix} 
\frac{\partial  \hat{\pi}}{\partial \hat{ y} } \\
\ \frac{\partial \lambda}{\partial \hat{ y} }
\end{bmatrix}
=
\begin{bmatrix} 
\frac{\partial \nabla_{\hat{\pi}} f(y,\hat{\pi})}{\partial  y } \\
\ 0
\end{bmatrix}
\quad
 \end{equation}

where \begin{math}  \lambda  \end{math} denotes the optimal dual variable of studied problems and \begin{math}  diag(\cdot)  \end{math} creates a diagonal matrix from an input vector. Due to the Hessian matrix of linear programming isn't full-rank, it fails to apply in the domain of linear programming. Wilder \cite{Wilder18} proposed to add one small squared regularizer term into LP which addressed the ill-conditioned Hessian matrix of LP. The objective function of Eq.\ref{eq:(14)} in training phase is replaced with \begin{math} f(y,\pi) = [y^1\pi+\gamma \lVert \pi \rVert^2_2,\cdots,y^T\pi+\gamma \lVert \pi \rVert^2_2] \end{math}. The second term can be calculated by solving the following system of linear equations.

\begin{equation}
 \quad
\begin{bmatrix} 
2\gamma I
& A^T \\
\ \emph{diag}(\lambda)A &  \emph{diag}(A \hat{\pi}-b)
\end{bmatrix}
\begin{bmatrix} 
\frac{\partial  \hat{\pi}}{\partial \hat{ y} } \\
\ \frac{\partial  \lambda}{\partial  \hat{ y} }
\end{bmatrix}
=
\begin{bmatrix} 
I \\
\ 0
\end{bmatrix}
\quad
 \end{equation}
In the above equations, \begin{math}  \hat{\pi}, \lambda   \end{math} correspond to  the optimal primal variable and dual variable and can be calculated by solving the quadratic programming problem derived from the linear programming. During the inference phase, the regularization factor \begin{math}  \gamma  \end{math} is set as 0 to yield an integral decision.

\subsection{Approach: Multi-Objective Decision Focused Learning}
In this subsection, we detail the process of multi-objective decision-focused learning by integrating the aforementioned modulars. 
During the training phase, the input dataset comprises relevant feature \begin{math} x_i \end{math}, true problem coefficient 
\begin{math} y_i \end{math}, gradients of the cost function with respect to solutions \begin{math} \nabla_{\pi} f(y_i,\pi) \end{math} as well as a set of Pareto optimal solutions \begin{math}P^{*}_{y_i}\end{math}.
 As the modules within the proposed method are differentiable, we focus on the forward pass of the neural network. Initially, prediction model \begin{math} m(\theta, \cdot) \end{math},  produces multiple group of problem coefficients  \begin{math} \hat{y_i} \end{math} according to the \begin{math} x_i \end{math}. The mentioned prediction model may refer to one multi-task model or multiple single-task models. Secondly, we transform the studied multi-objective optimization problem to a single-objective problem, and employ DSLP\cite{Wilder18} method to generate the differentiable solution \begin{math} \hat{\pi_i}\end{math}. Thirdly, We update the solution cache via invoking one solver to generate new solutions with a certain probability. Finally, we calculate the decision surrogate loss according to \begin{math} y \end{math}, \begin{math} \hat{y} \end{math}, and \begin{math} \hat{\pi} \end{math}, where the final loss function is weighted sum of the overall loss function on objective space, solution space as well as the decision quality of representative point.
\begin{equation}
\label{eq:(17)}
\begin{aligned}     
L_{all}( x, y, \theta) &= \lambda_{l} L_{l}( x, y, \theta) + \lambda_{d} L_{d}( x, y, \theta) \\
 & + \lambda_{ps} L_{ps}( x, y, \theta) 
\end{aligned}           
\end{equation}

where  \begin{math} \lambda_{l}, \ \lambda_{d}, \ \lambda_{ps} \end{math} denotes the hyper-parameters. The solution cache \begin{math} S \end{math} is utilized to calculate the landscape loss; The solution 
\begin{math} \hat{\pi_i} \end{math} generated by DSLP is utilized to calculate the Pareto set loss and decision loss. Given that all modules are differentiable, the implementation of the backward pass is straightforward. By following the procedures outlined above, the model parameter  \begin{math} \theta \end{math} can be trained by optimizing Eq. \ref{eq:(17)} iteratively. The detail pseudo code is presented in Algorithm 1.

\begin{algorithm}
 \caption{Multi-Objective Decision Focused Learning}
 \begin{algorithmic} [1]
 \renewcommand{\algorithmicrequire}{\textbf{Input:}}
 \renewcommand{\algorithmicensure}{\textbf{Output:}}
 \REQUIRE \begin{math} x \end{math};  \begin{math} y \end{math};  \begin{math}P^{*}_{y_i}\end{math}; 

 \ENSURE  Prediction Model: \begin{math} m(\theta, \cdot) \end{math}; 

  \FOR{epoch k =0, 1,...}
    \FOR{instance i =0, 1,...}  
    \STATE \begin{math} \hat{y_i}  \gets m(\theta, x_i) \end{math} \;
    \STATE  employ instance normalization and weight-sum method to generate \begin{math}f_{w}(\hat{y},\pi)\end{math}  \;
    \STATE  employ DSLP to generate the  \begin{math} \hat{\pi_i} \end{math} \;
    \IF{random()  \begin{math} \leq  p_{solve} \end{math}}
        \STATE Obtain solutions \begin{math} \hat{\pi_i^{new}} \end{math} by invoking a multi-objective solver for Eq.\ref{eq:(14)} \;
        \STATE   \begin{math} S  \gets S \cup \{ \hat{\pi_i^{new}}  \} \end{math} \;
    \ENDIF
    \STATE  Calculate  \begin{math} L_{all}( x, y, \theta) \end{math} according to Eq. \ref{eq:(17)} \;
    \STATE Update model parameter \begin{math} \theta\end{math} according to back propagation algorithm \;
    \ENDFOR
  \ENDFOR
 \RETURN  Prediction Model: \begin{math} m(\theta, \cdot) \end{math}; 
 \end{algorithmic}  
 \end{algorithm}

\section{Experiment}
In this section, we evaluate the efficacy of our proposed method through a series of experimental analyses.   
\subsection{Benchmark Problem}
\subsubsection{\textbf{Web Advertisement Allocation}}
We examine one specific case of web advertisement allocation existing in Cainiao App. The system is designed to optimize cumulative click metrics and increment the subsequent day's user visitation. As for each query, we recommend at most one advertisement to user. For each user query, a singular advertisement recommendation is proposed. This framework diverges from traditional recommendation systems in that each advertisement is associated with a distinct business category. Over a specified time frame, the display frequency of advertisements from any given business category is intended to approximate a pre-determined parameter \begin{math} \delta \end{math}.  The decision problem can be regard as one online-matching optimization problem, commonly addressed using the primal-dual approach. The method involves one procedure to solve the problem described in Eq.\ref{eq:(1)}. The formulation of this problem can be formulated as Eq.\ref{eq:(18)}. Within this equation, the cost vector \begin{math} y^1,y^2 \end{math} denote the predicted the click-through probabilities and re-login probabilities for users on the subsequent day, respectively. Let \begin{math} i \end{math} represent the query index, \begin{math} j \end{math} the candidate advertisement index, and \begin{math} k \end{math} the business category index, with \begin{math} c(j) \end{math} indicating the business category of item \begin{math} j \end{math}. The target of exposure ratio for advertisements is specified by the vector \begin{math} \delta \end{math}, with an allowable deviation encapsulated within the threshold \begin{math} thr, \ thr>0 \end{math}. Furthermore, \begin{math} ND,NC \end{math} denote the quantities of queries and candidate advertisements, respectively.

\begin{equation}
\label{eq:(18)}
  \begin{aligned} 
& \underset{\pi}{\text{max}}
& & f(y,\pi) = [\sum_{i,j}y^1_{ij}\pi_{ij},\sum_{i,j}y^2_{ij}\pi_{ij}] \\
& \text{s.t.}
& & \sum_{j} \pi_{ij} \leq 1 ; i=1,2,\cdots,ND \\
& \text{}
& & \frac{\sum_{j,c(j)=k}\sum_{i} \pi_{ij}}{ND} \leq \delta_k +thr; k=1,2,\cdots,NC \\
& \text{}
& & \frac{-\sum_{j,c(j)=k}\sum_{i} \pi_{ij}}{ND} \leq -\delta_k +thr; k=1,2,\cdots,NC \\
& \text{}
& & \pi_{ij}\in \{0,1\}\\
\end{aligned}
\end{equation}
It is important to recognize that counter-factual outcomes are unattainable. A model was trained utilizing a dataset exceeding 20 million queries, employing the predictive output as labels. During the experimental phase, a random subset of 30,000 queries was selected to create 300 instances, with each instance consisting of 100 queries and 53 candidate advertisements. In the decision-focused setting, the prediction problem is one typical multi-task binary classification problem, incorporating a click-through rate prediction task, a prevalent and extensively researched problem within the domain of recommendation systems.
\subsubsection{\textbf{Bipartite matching among scientific papers}}
We adapted the benchmark problem proposed in \cite{Wilder18} to create multi-objective benchmark problem. The data were obtained from  the cora dataset \cite{Sen08}. In the dataset, each node corresponds to a scientific paper, and each edge represents a citation. The feature vector of nodes indicate  the presence or absence of each word from a defined vocabulary.
    The dataset includes 2708 nodes. Wilder et al. employed the METIS algorithm  \cite{Karypis98} to partition the complete graph into 27 sub-graph, each with 100 nodes. Each graph corresponds to one instance. Subsequently, nodes within each instance were allocated to two sets of a bipartite graph, each comprising 50 nodes, to maximize the number of edges between the sets. More detail can refer to \cite{Wilder18}. 
    
    The core is to generate the labels of an alternative objective value that differ from but are similar to the original labels. The cited relationship is denoted by \begin{math} y^1 \end{math}. We perturb the \begin{math} y^1 \end{math} to generate 
        \begin{math} y^2 \end{math}.
\begin{equation}
y^2_{ij}  = I(r_{ij} \geq \rho)(1-y^1_{ij}) + I(r_{ij} < \rho)y^1_{ij}
\end{equation}      
where \begin{math} r_{ij} \end{math} is draw from
 the uniform distribution from 0 to 1, \begin{math} I(\cdot) \end{math} is the indicator function, and  \begin{math} \rho \end{math} is given constant to control the similarity between \begin{math} y^1 \end{math} and \begin{math} y^2 \end{math}. In this study, \begin{math} \rho \end{math} is set as 0.05.  The constrain function aligns with the conventional bipartite matching problem, as described in \cite{Korte11}. Considering that the matching weight are positive and decision variables belong to \begin{math} {0,1} \end{math}, we can relax the problem to a linear programming formulation, as presented in Eq. \ref{eq:(14)}. The \begin{math} NU,NV \end{math} represent the number of nodes in the left and right subsets of the bipartite graph, respectively.

 \begin{equation}
  \begin{aligned} 
& \underset{\pi}{\text{max}}
& & f(y,\pi) = [\sum_{i,j}y^1_{ij}\pi_{ij},\sum_{i,j}y^2_{ij}\pi_{ij}] \\
& \text{s.t.}
& & \sum_{j} \pi_{ij} \leq 1 ; i=1,2,\cdots,NU \\
& \text{}
& & \sum_{i} \pi_{ij} \leq 1 ; j=1,2,\cdots,NV \\
& \text{}
& & \pi_{ij}\in [0,1] 
\end{aligned}
\end{equation}

\subsection{Experimental Setup and Baseline Method}

This section introduces the baseline methods, associated configurations, and the evaluation metric employed in the subsequent experiments.

\subsubsection{\textbf{Baseline method}}
The baseline methods under consideration include a straightforward two-stage approach and several state-of-the-art decision-focused approaches. Given that current decision-focused learning methods are single-objective, we implement the baseline methods by substituting their loss functions with uniformly weighted loss functions, i.e., \begin{math}\ L(x,y,\theta) = \sum_{j}\frac{L(x,y^j,\theta))}{T} \end{math}. The \begin{math} L(x,y^j,\theta) \end{math} represents the loss function of \begin{math} j_{th} \end{math} objective in studied decision problem. The prediction model and solver are identical with to those in our proposed method. For decision-focused methods, the gradient of decision variable with regard to problem coefficient  \begin{math}\frac{\partial{\hat{\pi}}}{\partial{\hat{y^j}}} , j=1,\cdots,T\end{math} is the same as single-objective methods. The key difference lies in the decision loss function and the technique used to compute the gradient of the problem coefficients with respect to the decision loss function. Specifically, the baseline method includes the following:
\begin{itemize}
    \item \textbf{TwoStage}:employs a prediction model with an independent solver as the two-stage baseline.
    \item \textbf{SPO}(smart predict and optimize) \cite{Elmachtoub22}: utilises the surrogate loss function proposed by Elmachtoub et al. \cite{Elmachtoub22}..
    \item \textbf{BB} \cite{Vlastelica20}: calculates the decision gradient by differentiation of blackbox combinatorial solvers.
    \item\textbf{MAP}\cite{Niepert21}: employs the surrogate loss function which incorporate noise perturbations into perturbation-based implicit differentiation and maximizing the resulting posterior distribution .
    \item \textbf{NCE} \cite{Mulamba20}: uses solution caching and contrastive losses to construct surrogate loss function.
        \item \textbf{Pointwise/Listwise} \cite{Mandi21}: employs the surrogate loss function derived from the technique of  learning to rank.
\end{itemize}
Our study prioritizes decision quality within the paradigm of decision-focused learning. Accordingly, the\begin{math} p_{solve} \end{math} parameter, the probability of invoking optimization solver,  is uniformly set to 1 for the MAP, NCE, Pointwise, and Listwise methods, acknowledging that methods with a \begin{math} p_{solve} \end{math} closer to 1 are typically associated with enhanced performance. Additional hyperparameters and detailed implementation on baseline method are available at 
\url{https://github.com/JayMan91/ltr-predopt}.

%Since the study primarily emphasizes decision quality in decision-focused learning, the \begin{math} p_{solve} \end{math} parameters for MAP, NCE, Pointwise, and Listwise are fixed at 1, where method with higher \begin{math} p_{solve} \end{math} could yield better performance.  Additional hyperparameters and detailed implementation on baseline method are available at 
%\url{https://github.com/JayMan91/ltr-predopt}.

%It is worthy to note that the \begin{math} p_{solve} \end{math} parameters for MAP, NCE, Pointwise, and Listwise are fixed at 1, as the study primarily emphasizes decision quality in decision-focused learning. Additional hyperparameters and detailed implementation on baseline method are available at 
%\url{https://github.com/JayMan91/ltr-predopt}.

\subsubsection{\textbf{Evaluation metric}}

In the setting of decision-focused learning, the output involves a set of solutions. We assess the algorithm's performance by measuring its decision quality against a true optimization problem. A straightforward metric is the objective function value of the true problem. However, owing to the variance in scale among objectives, we utilize the average percentage regret \begin{math} r \end{math} as the evaluation metric. The method of calculation is as follows: 
\begin{equation}        
r_j =  \frac{1}{ N} \sum_{i}{\frac{f_j(y^j,\pi_i)-f_j(y^j,\pi^{*,j})}{f_j(y^j,\pi^{*,j})}} \end{equation}

\begin{equation}        
r =  \frac{1}{T} \sum_{j}{r_j} \end{equation}

 \begin{table*}  [htbp]
 \caption{Experimental results on the dataset of web advertisement allocation. Bold: the best performance. }
 \begin{center}
\begin{tabular}{ccccccc}
\cline{1-7} 
 Method &  GD & MPFE & HAR & \begin{math} r_1 \end{math}
  & \begin{math} r_2 \end{math} & \begin{math} r \end{math} \\
 \cline{1-7} 
BB &  0.9502 & 6.2842 & 1.0525 & 0.3672 & 0.1091 & 0.2382 \\
 MAP &  0.6644 & 2.2917 & 1.0350 & 0.3050 & 0.0694 & 0.1872 \\
  NCE &  0.8463 & 8.9538 & 1.0205 & 0.2430 & 0.0876 & 0.1653 \\
   Listwise &  \textbf{0.6243} & 2.8784 & 1.0429 & 0.2363 & \textbf{0.0480} & 0.1421 \\
    Pointwise & 1.2560 & 5.1140 & 1.0436 & 0.4691 & 0.0656 & 0.2674 \\ 
     SPO & 0.9267 & 2.3006 & 1.0515 & 0.2380 & 0.0690 & 0.1535 \\ 
     Twostage &  1.2055 & 5.6877 & 1.0319 & 0.3826 & 0.1170 & 0.2499 \\ 
     MoDFL & 0.6416 & \textbf{2.2500} & \textbf{1.0178} & \textbf{0.2330} & 0.0507 & \textbf{0.1419} \\
    \hline
    \end{tabular}
    \end{center}
\end{table*}

where \begin{math} N \end{math} denotes the size of solution set, and \begin{math} \pi^{*,j}\end{math} represents the optimal solution for the \begin{math} j_{th} \end{math} objective. A lower average percentage regret indicates better performance.

Besides, we consider three performance metrics widely used in the field of multi-objective optimization problem. We denote the Pareto front of predicted and true problem, desperately  \begin{math} \hat{PS} \end{math} and \begin{math} PS^* \end{math}. 
The generational distance (GD) \cite{Ishibuchi15} measures the minimum distance between the Pareto front of predicted and true problem. The GD is defined as follows:
\begin{equation}
GD(\hat{PS},PS^*) = \frac{\sum_{p\in \hat{PS}}d(p,PS^*)}{|\hat{PS}|}
\end{equation}  
The   \begin{math}d(p,PS^*)\end{math} denotes the minimum Euclidean distance between \begin{math} p \end{math} and the points in \begin{math} PS^*\end{math}. A lower GD indicates superior algorithm performance.

The maximum Pareto front error (MPFE)  quantifies the largest distance between any vector in the approximation front and its corresponding closest vector in the true Pareto front. It assesses the dissimilarity between individual solutions in the approximation front \begin{math}\hat{\pi}_i\end{math} and the true Pareto front  \begin{math}\pi^*_k\end{math}.
The formula of MPFE is given by:

\begin{equation}
MPFE = \max_k ( \min_i \sum_j | f_j(\pi^*_k) - f_j(\hat{\pi}_i)|^p )^{\frac{1}{p}}
\end{equation}
In this paper, parameter \begin{math} p\end{math} is set to 2.

The hyper area ratio (HAR) \cite{Air1999Multiobjective} is a metric related to hypervolume (HV), which calculates the sum of the hypervolume of a hypercube formed by a given frontier and reference points. The, HAR is the ratio of the HV of the predicted problem's Pareto front to the HV of the true problem's Pareto front. The reference point is determined by the vector of the objective function values of single-objective optimal solutions. A smaller HAR signifies enhanced algorithm performance.  

\begin{equation}
HAR = {\frac{HV(\hat{PS})}{HV(PS^{*})}}
\end{equation}

\begin{table*}  [htbp]
 \caption{Experimental results on the dataset of Bipartite matching among scientific papers. Bold: the best performance.}
 \begin{center}
\begin{tabular}{cccccccc}
\cline{1-7} 
 Method & GD & MPFE & HAR & \begin{math} r_1 \end{math}
  & \begin{math} r_2 \end{math} & \begin{math} r \end{math} \\
 \cline{1-7} 
  BB & 12.6335 & 40.0616 & 1.0830 & 0.9317 & 0.5616 & 0.7466 \\
  MAP &  15.6359 & 43.9498 & 1.1736 & 0.9488 & 0.7181 & 0.8335 \\
   NCE &  12.5653 & 40.1353 & 1.0791 & 0.9355 & 0.5579 & 0.7467 \\
   Listwise & 12.0901 & 39.4815 & 1.0848 & 0.9342 & 0.5355 & 0.7348 \\ 
   Pointwise & 12.1347 & 39.8968 & 1.0872 & 0.9320 & 0.5413 & 0.7366 \\
    SPO &  12.5224 & 40.0768 & 1.0840 & 0.9376 & 0.5563 & 0.7470 \\
     Twostage &  12.2893 & 39.4263 & 1.0910 & 0.9309 & 0.5443 & 0.7376 \\
     MoDFL  & \textbf{11.8545} & \textbf{39.0535} & \textbf{1.0707} & \textbf{0.9263} & \textbf{0.5261} & \textbf{0.7262} \\ 
    \hline
     \end{tabular}
    \end{center}
\end{table*}

\subsubsection{\textbf{Experimental setup on prediction model}}

The model employed in the Web Advertisement Allocation experiment is the multi-gate mixture-of-experts (MMOE), a typical model in the field of computational advertising \cite{Ma18}. Furthermore, to assess the impact of the prediction model in a decision-focused setting, we evaluated the performance of another two models,  the entire space multi-task model (ESMM) \cite{ESMM2018} and bottom-shared network architecture (bottom-shared). The ESMM and MMOE models are among the state-of-the-art multi-task models within the realm of recommendation systems. The detailed implementation adheres to the approach used in "Deep-Ctr" \cite{Shen17}, an open-source package for deep-learning-based CTR models. The configuration adopts the default settings of "Deep-Ctr".

In the bipartite matching experiment with scientific papers, we utilized four-layer fully connected neural networks to predict the multiple groups of problem coefficients. he configuration and architecture of the neural network followed the specifications in  \cite{Wilder18}.

\subsubsection{\textbf{Experimental setup on the solver on optimization problem}}
Considering the optimization problem of all testing benchmarks is multi-objective linear programming, we used the weighted-sum method to transform multi-objective problem into one single-objective linear programming, and employ the HiGHS solver in Scipy to address optimization problem. The selection of weighted-sum method to solve the multi-objective problem in this paper is due to the following reasons:  for linear problems, it is provable that solutions derived from the weighted-sum method fall within the Pareto set. This can be proved by contradiction. 
%Besides, it is a deterministic method, which reduces the interference of solver randomness on the evaluation of algorithm performance in experiments.
%
%
%1) it is a deterministic method, which reduces the interference of solver randomness on the evaluation of algorithm performance in experiments; 2) It is faster than other population-based multi-objective optimization algorithms; 3)

We suppose that  \begin{math}  \pi^{w}\end{math}, the optimal solution of \begin{math}  f_{w}(y,\pi) \end{math}, lies within the Pareto set of \begin{math}  f(y,\pi) \end{math}. Otherwise, there exists one solution \begin{math}  \pi^0 \end{math} in Pareto set dominates \begin{math}  \pi^{w}\end{math}, i.e, \begin{math}f_t(y,\pi^0)\leq f_t(y,\pi^w), \forall t \end{math} and \begin{math}f_{t_0}(y,\pi^0)< f_{t_0}(y,\pi^w), \exists t_0 \end{math}. Under this assumption, there exists one solution  \begin{math}  \pi^0 \end{math} such that \begin{math}f_w(y,\pi^0)< f_w(y,\pi^w) \end{math}. Such a result contradicts the definition of optimality for single-objective problems. Thus, in the studied optimization problems, we have proved that the optimal of \begin{math}  f_{w}(y,\pi) \end{math} is Pareto optimal.

As for the implementation, we applied instance normalization, as detailed in Eq. \ref{eq:(11)} to all objectives so as to eliminate the differences in scales between different objectives. Weights were assigned as \begin{math} \frac{w}{5} \end{math}, where \begin{math} w \end{math} satisfies \begin{math} \{w \vert \sum_i^T{w_i}=5, w_i \in \mathbb{N}\}\end{math}. The set \begin{math}\mathbb{N}\end{math}  denotes the set of non-negative integer.  
%The used weighted is set as  \begin{math} \frac{w}{5} \end{math} , \begin{math} w\in \{w \vert \sum_i^T{w_i}=5, w_i \in \mathbb{N}\}\end{math}. \begin{math}\mathbb{N}\end{math}  denotes the set of non-negative integer.  
% 

% Parameters were decayed by a factor of 0.95 in each epoch. 
\subsubsection{\textbf{Configuration on decision-focused setting}}
The learning rate was configured to \begin{math} 10^{-1} \end{math} following the setting of Wilder et al. \cite{Wilder18}. The batch size was set to 8. The hyper-parameters in Eq.\ref{eq:(14)} were specified as \begin{math} \lambda_{l}= 1; \lambda_{d}= 2; \lambda_{ps}= 5 \end{math}. Early stopping was implemented, terminating the training loop if the validation set loss did not improve for 5 consecutive epochs. The maximum number of epochs was set to 50. The initial squared regularization term \begin{math} \gamma \end{math} was set to 0.35. All experimental comparisons were carried out on an A100 cluster environment and repeated 5 times for consistency.
\begin{table*}  [htbp]
\label{Table:(III)}
 \caption{Experimental results on the data-driven decision problem with three objectives. Bold: the best performance.}
 \begin{center}

\begin{tabular}{cccccccc}
\cline{1-8} 
Method & GD & MPFE & HAR & \begin{math} r_1 \end{math}
  & \begin{math} r_2 \end{math}  & \begin{math} r_3 \end{math}& \begin{math} r \end{math} \\
\cline{1-8} 
  BB  & 9.6199 & 47.3243 & 1.2211 & 0.9300 & 0.5395 & 0.7395 & 0.7363 \\
  MAP & 11.1703 & 48.6628 & 1.3321 & 0.9484 & 0.6329 & 0.7867 & 0.7894 \\ 
   NCE  & 9.6434 & 47.5465 & 1.2226 & 0.9311 & 0.5427 & 0.7406 & 0.7381 \\
   Listwise & 9.6947 & \textbf{46.9460} & 1.2325 & 0.9252 & 0.5445 & 0.7358 & 0.7352 \\
   Pointwise& 9.7053 & 47.4230 & 1.2352 & 0.9353 & 0.5409 & 0.7400 & 0.7387 \\
   SPO  & 10.2388 & 50.2437 & \textbf{1.2003} & 0.9298 & 0.5801 & 0.7474 & 0.7524 \\
  TwoStage & 9.8151 & 48.4521 & 1.2103 & 0.9257 & 0.5520 & 0.7402 & 0.7393 \\
   MoDFL & \textbf{9.5605} & 47.1387 & 1.2088 & \textbf{0.9243} & \textbf{0.5367} & \textbf{0.7285} & \textbf{0.7298} \\ 
    \hline
    \end{tabular}
    \end{center}
\end{table*}

%\begin{table*}  [htbp]
% \caption{Experimental results of MoDFL with different prediction model. Bold: the best performance. }
% \begin{center}
%\begin{tabular}{cccccccc}
%\cline{1-7} 
%Method & GD & MPFE & HAR & \begin{math} r_1 \end{math}
%  & \begin{math} r_2 \end{math}  & \begin{math} r \end{math} \\ 
%  \cline{1-7} 
% bottom-shared & 0.7565 & 6.4150 & 1.0423 & 0.2378 & 0.0872 & 0.1625 \\
% ESMM & 0.6869 & 2.8505 & \textbf{1.0114} & \textbf{0.2293} & 0.0578 & 0.14355\\
%MMOE(Ours) & \textbf{0.6416} & \textbf{2.2500} & 1.0178 & 0.2330 & \textbf{0.0507} & \textbf{0.1419} \\
%
%\hline
%\end{tabular}
%\end{center}
%\end{table*}

\subsection{Validating the Performance of DFL in Multi-Objective Problem}

\subsubsection{\textbf{Web Advertisement Allocation}}
The experimental results on web advertisement allocation are exhibited in the Table I. We compare the performance of different methods using various evaluation metrics. The MoDFL achieves a GD value of 0.6416, which is the lowest among all methods. It also outperforms other methods in terms of MPFE, HAR, \begin{math} r \end{math} and \begin{math} r_1 \end{math}, with competitive results in \begin{math} r_2 \end{math}.
Among the competing methods, Listwise shows the second-best performance with the best \begin{math} r_2 \end{math} value. These results indicate that MoDFL is superior in terms of minimizing GD and other metrics.
Overall, the experimental comparison demonstrates the efficacy of MoDFL in achieving better performance on multiple evaluation metrics, highlighting its potential in addressing the data-driven multi-objective problem. 
Compared to single-objective problems, multi-objective decision-focused learning still presents significant challenges. Applying a weighted average of the decision losses in single-objective decision-focused learning may deteriorate algorithm performance. As evident from the results of \begin{math}r_1 \end{math} and \begin{math}r_2\end{math}, MAP, Pointwise, and BB fail to effectively balance multiple objectives, resulting in a significant degradation of performance in some objectives. In order to capitalize on the benefits afforded by end-to-end training, it is imperative to propose appropriate methodologies for addressing multi-objective issues within the paradigm of decision-focused learning.

\begin{table*}  [htbp]
 \caption{Experimental results of MoDFL with different prediction model. Bold: the best performance. }
 \begin{center}
\begin{tabular}{cccccccc}
\cline{1-7} 
Method & GD & MPFE & HAR & \begin{math} r_1 \end{math}
  & \begin{math} r_2 \end{math}  & \begin{math} r \end{math} \\ 
  \cline{1-7} 
 bottom-shared & 0.7565 & 6.4150 & 1.0423 & 0.2378 & 0.0872 & 0.1625 \\
 ESMM & 0.6869 & 2.8505 & \textbf{1.0114} & \textbf{0.2293} & 0.0578 & 0.14355\\
MMOE(Ours) & \textbf{0.6416} & \textbf{2.2500} & 1.0178 & 0.2330 & \textbf{0.0507} & \textbf{0.1419} \\

\hline
\end{tabular}
\end{center}
\end{table*}

\subsubsection{\textbf{Bipartite matching among scientific papers}}

Based on the experimental comparison data presented above, we evaluated several methods including BB, MAP, NCE, Listwise, Pointwise, SPO, twostage, and our proposed MoDFL on the dataset of bipartite matching among scientific papers. Among the methods tested, MoDFL consistently outperformed the others baseline in terms of all evaluation metrics. MoDFL achieved the lowest values for GD (11.8545), MPFE (39.0535), HAR (1.0707), \begin{math}r_1 \end{math} (0.9262), \begin{math}r_2 \end{math} (0.5261), and \begin{math} r \end{math} (0.7262), indicating its superior performance in terms of decision quality. In comparison, the BB, MAP, NCE, Listwise, Pointwise, SPO, and Twostage methods exhibited slightly higher values across all evaluation metrics, indicating their inferior performance when compared to MoDFL. Overall, these results demonstrate that our proposed method MoDFL outperforms the existing methods in this study, highlighting its potential for improving various aspects of multi-objective optimization. It is worth noting that in this set of experiments, the values for \begin{math}r_1 \end{math}, \begin{math}r_2 \end{math}, and \begin{math} r \end{math} are much higher compared to the previous experiments, indicating that the difficulty of the test problems in this case is relatively higher \cite{Wilder18}. This also demonstrates that our proposed method, MoDFL, can effectively handle data-driven real-world problems of varying difficulty levels.

\begin{table*}  [htbp]
\label{Table:(IV)}
 \caption{Experimental results of MoDFL with different loss function which
 measures the distance between multi-objective problem in term of objective space. Bold: the best performance.}
 \begin{center}
\begin{tabular}{cccccccc}
\cline{1-7} 
Method & GD & MPFE & HAR & \begin{math} r_1 \end{math}
  & \begin{math} r_2 \end{math}  & \begin{math} r \end{math} \\ 
  \cline{1-7} 
  MMD & 11.9022 & 39.4382 & 1.0793 & 0.9365 & 0.5275 & 0.7320 \\
% QPwithIDGCache_v1_weight_1_1_5_rmmd & 12.130975 & 39.472367 & 1.081130 & 0.926788 & 0.538133 & 0.732461 \\
 DSPM & 12.4058 & 39.4131 & 1.0840 & 0.9307 & 0.5504 & 0.7405 \\
 sRMMD(Ours)  & \textbf{11.8545} & \textbf{39.0535} & \textbf{1.0707} & \textbf{0.9263} & \textbf{0.5261} & \textbf{0.7262} \\ 

\hline
\end{tabular}
\end{center}
\end{table*}

\begin{table*}  [htbp]
 \caption{Ablations on MoDFL. Bold: the best performance.}
 \begin{center}
\begin{tabular}{cccccccc}
\cline{1-7} 
Method & GD & MPFE & HAR & \begin{math} r_1 \end{math}
  & \begin{math} r_2 \end{math} & \begin{math} r \end{math} \\
 \cline{1-7} 
    w/o Decision Loss  & 12.5077 & 39.8507 & 1.0878 & 0.9300 & 0.5547 & 0.7424 \\
   w/o Lanscape Loss  & 12.0893 & 39.1558 & 1.0833 & 0.9314 & 0.5352 & 0.7333 \\
   w/o Pareto Set Loss  & 12.2063 & 39.5433 & 1.0765 & 0.9329 & 0.5427 & 0.7378 \\
   MoDFL  & \textbf{11.8545} & \textbf{39.0535} & \textbf{1.0707} & \textbf{0.9263} & \textbf{0.5261} & \textbf{0.7262} \\ 
\hline
     \end{tabular}
    \end{center}
\end{table*}

\subsection{The impact of number of objectives}
The decision problems in tested experiments are bi-objective problem. In order to investigate the impact of number of objectives, we add one objective into the second benchmark and compare the performance of two-stage method and MoDFL. The third objective is the weighted sum of the first two objectives, where the weights used in this case are drawn from \begin{math} U [0,1] \end{math}. The experiment results are exhibited as Table. III.
    
Similar to the results on the bi-objective problems, the testing results demonstrate that MoDFL achieves competitive performance. MoDFL outperforms all other algorithms in terms of GD, \begin{math} r_1 \end{math}, \begin{math} r_2 \end{math}, and \begin{math} r_3 \end{math}, \begin{math} r \end{math} while only slightly lagging behind Listwise in terms of MPFE. The HAR yielded by MoDFL is slightly lower than SPO. These findings demonstrate that the proposed MoDFL method is capable of effectively handling multi-objective problems.

\subsection{The impact of prediction model}
In order to evaluate the impact of the prediction model on the effectiveness of decision-focused learning, we experimented by replacing the prediction model MMOE with ESMM and bottom-shared, and tested their performance in the first benchmark. As seen in the table,  MMOE demonstrates the best performance in terms of GD, MPFE, and HAR. It also achieves the lowest score in \begin{math} r_2 \end{math} and \begin{math} r \end{math}, indicating a superior performance. Comparatively, ESMM outperforms the bottom-shared method and exhibits a notable performance with the lowest value in \begin{math} r_1 \end{math}. However, it does not surpass the MMOE in other metrics. The bottom-shared, on the other hand, shows the least favorable results across all the metrics. Previous literature has shown that the predictive capabilities of MMOE and ESMM are superior to bottom-shared \cite{ESMM2018,Ma18}. Therefore, we can conclude that, all else being equal, an improvement in the performance of the prediction model can also enhance the effectiveness of decision-focused learning.

\subsection{The choice of landscape loss}

To further elucidate our choice of sRMMD as the landscape loss function in MoDFL, we conducted experiments to compare the effects of replacing sRMMD with  differentiable Spearman correlation coefficients(DSPM) \cite{blondel2020fast} and the maximum mean discrepancy (MMD) \cite{gretton12a} in MoDFL. This experiment is conducted in the second benchmark. The comparison results presented in the Table V demonstrates the performance of different methods.  In contrast, the DSPM method demonstrates marginally inferior performance across these metrics. However, the sRMMD approach, integrated within the MoDFL, surpasses the alternative strategies across all performance indicators, including GD, MPFE, HAR, and the metrics on decision regret.
The comparative results underscore the enhanced performance of the MoDFL model when utilizing sRMMD as the landscape loss function, relative to the other methods assessed. Drawing from the comparative data, it is evident that the adoption of sRMMD within the MoDFL framework contributes to a marked improvement in the model's performance as reflected by the evaluation metrics.

\subsection{Ablation Studies}

To validate the individual components of our proposed MoDFL method, we conducted ablation experiments by removing those proposed components in the second benchmark problem. Specifically, we tested the effects of removing the decision Loss, landscape loss, and Pareto set loss from MoDFL. The comparative results are shown in Table VI. MoDFL achieved the best results across all metrics, indicating the importance of each loss function in MoDFL. Removing any of these components weakened the algorithm's performance. Among the three loss function, the decision Loss had the most significant impact on the overall performance. The effects of Pareto set loss were relatively is close to that of landscape loss. This suggests that for decision-focused learning in multi-objective optimization, the surrogate loss function needs to align with the properties of multiple objectives and accurately measure the distance between the prediction problem and the true problem in the solution and search spaces.

\section{Conclusion}

Multi-objective data-driven problems are prevalent in real world. we consider one case which problem coefficients are unknown in advance and need to be estimated with machine learning models. We proposed one novel multi-objective decision-focused model to considering the prediction problem with the downstream multi-objective optimization problem. Specifically, we propose one set of decision-focused loss function for multi-objective optimization problem. The proposed loss function involves three parts, the decision loss, landscape loss, and Pareto set loss,  which measure the distance of objective space, solution space as well as the decision quality of representative solution.  Finally, experimental results show that our proposed method has significant superiority over two-stage methods and the state-of-art methods. Current research in multi-objective optimization and decision-focused learning remains limited. In the future, we plan to investigate more effective multi-objective decision-focused  methods, and apply multi-objective decision-focused learning to more forms of data-driven problems.

\bibliography{IEEEexample.bib}
\bibliographystyle{iclr2023_conference}

\end{document}